\documentclass[letterpaper, 10 pt, conference]{ieeeconf}

\usepackage{amsmath}
\usepackage{amssymb}
\usepackage{mathtools}

\usepackage{tabularx}
\usepackage{graphicx}
\usepackage{bm}
\usepackage{color}
\usepackage{float}
\usepackage{units}
\usepackage{url}
\usepackage[hidelinks]{hyperref}
\usepackage{balance}
\usepackage{amsmath}
\usepackage{makecell}
\usepackage{cite}
\usepackage{contour}
\usepackage[table,dvipsnames]{xcolor}
\usepackage{amsthm}
\usepackage{algorithm, algorithmicx, algpseudocode}
\usepackage{booktabs, multirow, multicol}
\usepackage{adjustbox}
\usepackage{empheq}

\definecolor{lightgrayrow}{RGB}{240,240,240}
\definecolor{lightpinkrow}{RGB}{255,230,235}

\newcommand{\newsec}[1]{\vspace{0.1cm} \noindent \textbf{#1} }

\makeatletter 
\def\endfigure{\end@float} 
\def\endtable{\end@float}
\makeatother

\usepackage{subcaption}
\usepackage[]{graphicx}
\usepackage{caption}

\usepackage{lipsum}
\usepackage{subcaption}

\IEEEoverridecommandlockouts

\begin{document} 

\title{\Large \bf PAC-MAN: Perception-Aware CBF-RL for Whole-Body Safety \\ in Humanoid Dodgeball}

\author{Lizhi Yang, Junheng Li and Aaron D. Ames
\thanks{California Institute of Technology, CA 91106, USA}\thanks{Corresponding author email:{\tt\small lzyang@caltech.edu}. This work is supported by The Dow Chemical Company project \#227027AW and in part by Technology Innovation Institute.}
\vspace{-1.2cm}												
}	
\maketitle

\begin{abstract}
We present PAC-MAN, a perception-aware CBF-RL framework that couples control-barrier safety with deployment-realistic onboard sensing for whole-body humanoid dodgeball.
The deployed policy sees the ball only as segmentation-masked depth from a head-mounted camera, while training-time CBF guidance represents clearance to every body link, and an adversarial motion prior regularizes the resulting evasive reflexes.
We evaluate on a controlled any-link contact benchmark with seeded throws in two regimes: single throws and a deployment loop in which the robot walks back to its station and recovers between throws.
On this benchmark, the policy comes within a few points of a privileged state oracle: a fixed onboard camera alone is adequate for evasion.
We find that usable barrier structure depends on perceptual observability: Joint-CBF gives the best performance with accurate ball states, degrades under fixed-camera observations when used only as training guidance, and recovers with a ball-tracking gimbal or privileged runtime filter.
We therefore deploy a lightweight Link-CBF policy zero-shot on the Unitree G1 in the real world, where it tolerates imperfect perception, succeeds on $\mathbf{95\%}$ of throws, and uses semantic segmentation to dodge different balls.
\end{abstract}

\section{Introduction}
\label{sec:introduction}

Humanoid robots operating around moving objects must react within a fraction of a second, and the evasion itself must not cost them their balance.
Dodgeball isolates this safety problem in a short-horizon, whole-body setting: the robot must perceive an incoming ball, determine which links are threatened, and coordinate the torso, arms, and legs quickly enough to avoid contact while remaining upright.
Humans solve the same problem by keeping their eyes on the ball and moving their whole body out of its path.
Recent learning-based controllers can produce dynamic humanoid locomotion~\cite{gu2026humanoid,zhuang2025humanoidparkour}, whole-body imitation~\cite{gmt2025,sonic2025,beyondmimic2025}, and visually conditioned skills~\cite{visualmimic2025,hao2026cref}, but reactive whole-body safety from onboard vision remains difficult~\cite{he2024agilebut,kohlbrenner2026egocentric}, especially for a humanoid.

\begin{figure}[t]
  \centering
  \includegraphics[width=\columnwidth]{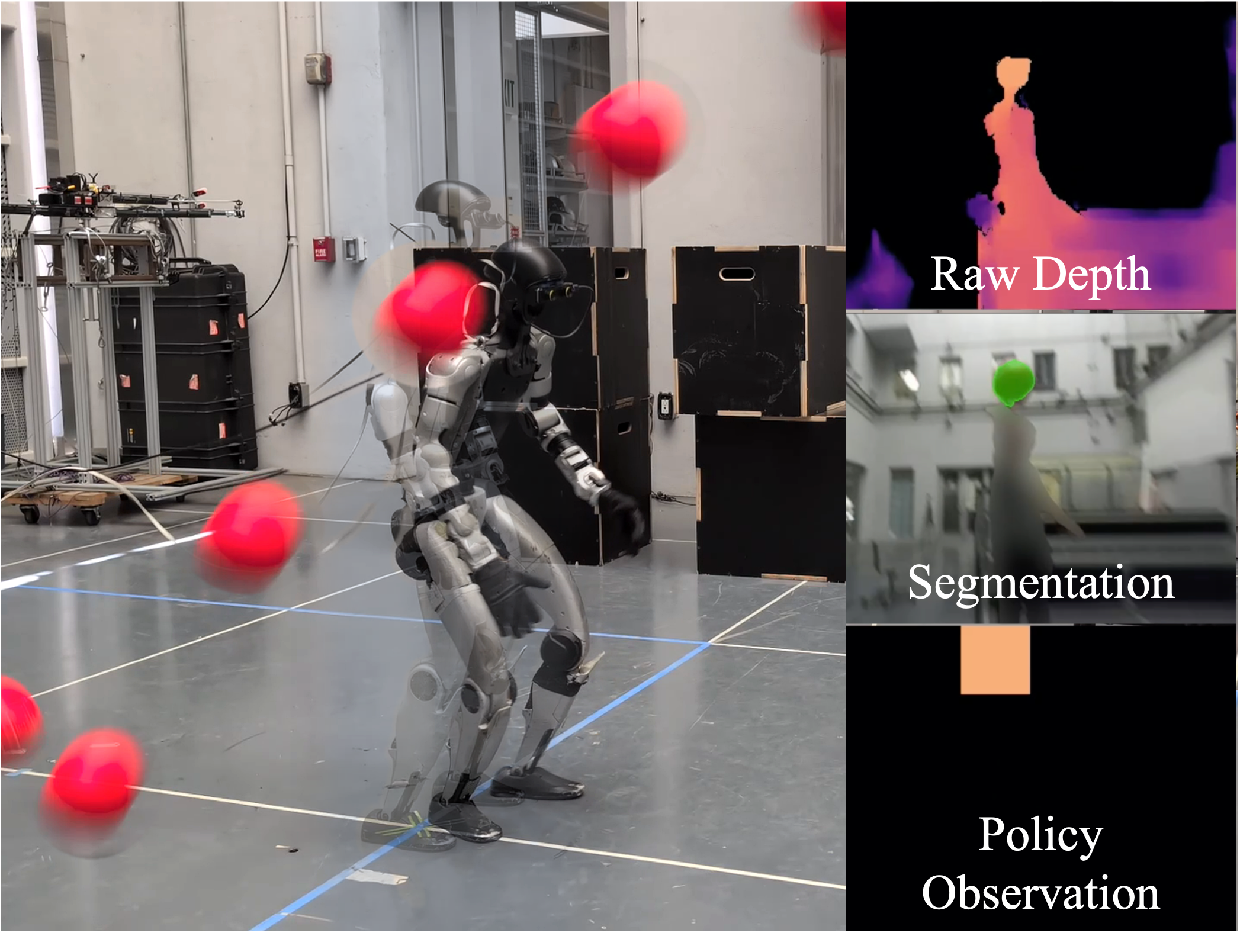}
  \caption{Perception-aware dodgeball couples safety and sensing: a humanoid must infer an incoming threat from a head-mounted depth camera, move every threatened link out of its path, and preserve balance without privileged ball states at deployment.}
  \label{fig:teaser}
\end{figure}

Control barrier functions (CBFs) encode collision-free sets and correct unsafe commands, provided the relevant state is available~\cite{ames2016control}.
Onboard perception breaks this convenient assumption: the incoming object may be visible only briefly, may leave the camera field of view, or may appear as a few noisy depth pixels.
CBF-RL can move barrier information into policy training~\cite{yang2025cbf}, but a deployed policy can internalize only as much safety structure as its observations support.
The threat representation and the barrier design therefore have to be treated as one coupled safety problem rather than as independent modules.

We study this coupling through perception-aware humanoid dodgeball (Fig.~\ref{fig:teaser}).
A ball is launched toward a humanoid robot, which must avoid contact with every body link and preserve balance using only proprioception and onboard depth at deployment.
Unlike navigation around a mapped obstacle, the hazard is transient, and the robot has no nominal velocity command or reference motion that specifies when and how to evade.
The short observation window and whole-body collision geometry make dodgeball a direct test of safety under partial observation.

We present PAC-MAN, a perception-aware CBF-RL framework for humanoid dodgeball.
The central question is how much barrier structure a policy can internalize when it receives only deployment-realistic observations.
PAC-MAN studies two levels: Link-CBF, a lightweight per-link reward used during training, and Joint-CBF, a stronger whole-body joint-space projection that can either guide training or stay active at test time as a privileged runtime filter.
The two levels separate learned barrier-guided behavior from online barrier enforcement.

How well the learned reflex works depends on what perception tells the policy about the threat.
The deployed system segments the ball in RGB and masks the depth image down to a compact ball-only observation.
An active gimbal, studied in simulation, tests whether keeping the threat in view for longer makes the stronger barrier structure usable.
An adversarial motion prior (AMP) regularizes how the humanoid moves, so the evasions come out as crouches, leans, sidesteps, and leaps; the safety structure itself comes from the CBF terms.

We use a controlled, contact-based benchmark to ask three questions: whether barrier guidance improves whole-body avoidance, how much barrier structure a policy can internalize without a runtime filter, and how the available perception changes the answer.
The benchmark calibrates on-target throws against a frozen statue and covers both single throws and a deployment loop in which the robot walks back to its station and recovers between throws.
Across a privileged state oracle, a fixed onboard camera, and an oracle-aimed active gimbal, we find that stronger barrier structure needs more informative observations.
The full Joint-CBF with runtime filtering is strongest when accurate ball states are available, whereas the lighter Link-CBF is the best deployable option under limited onboard perception.
We validate this design choice by deploying the fixed-camera Link-CBF policy zero-shot on a Unitree G1 using only onboard depth and proprioception.

\subsection{Contributions}
Our contributions are as follows:
\begin{itemize}
  \item we introduce perception-aware CBF-RL for whole-body humanoid reaction, pairing deployment-realistic observations with Link-CBF and Joint-CBF safe sets over every body link; the policy can internalize these structures during training, or a runtime filter can enforce them when state information permits.
  \item we evaluate under an any-link contact criterion and show that the appropriate barrier mechanism depends on the threat representation: Joint-CBF benefits from accurate ball state and oracle-aimed tracking, while Link-CBF remains effective with a fixed camera and no runtime filter.
  \item we deploy the fixed-camera Link-CBF policy zero-shot on the Unitree G1 using only onboard depth and proprioception; the robot succeeds on $95\%$ of throws, and semantic segmentation lets the same policy evade different balls.
  We will release the benchmark and training pipeline.
\end{itemize}

\begin{figure*}[t]
  \centering
  \includegraphics[width=\textwidth]{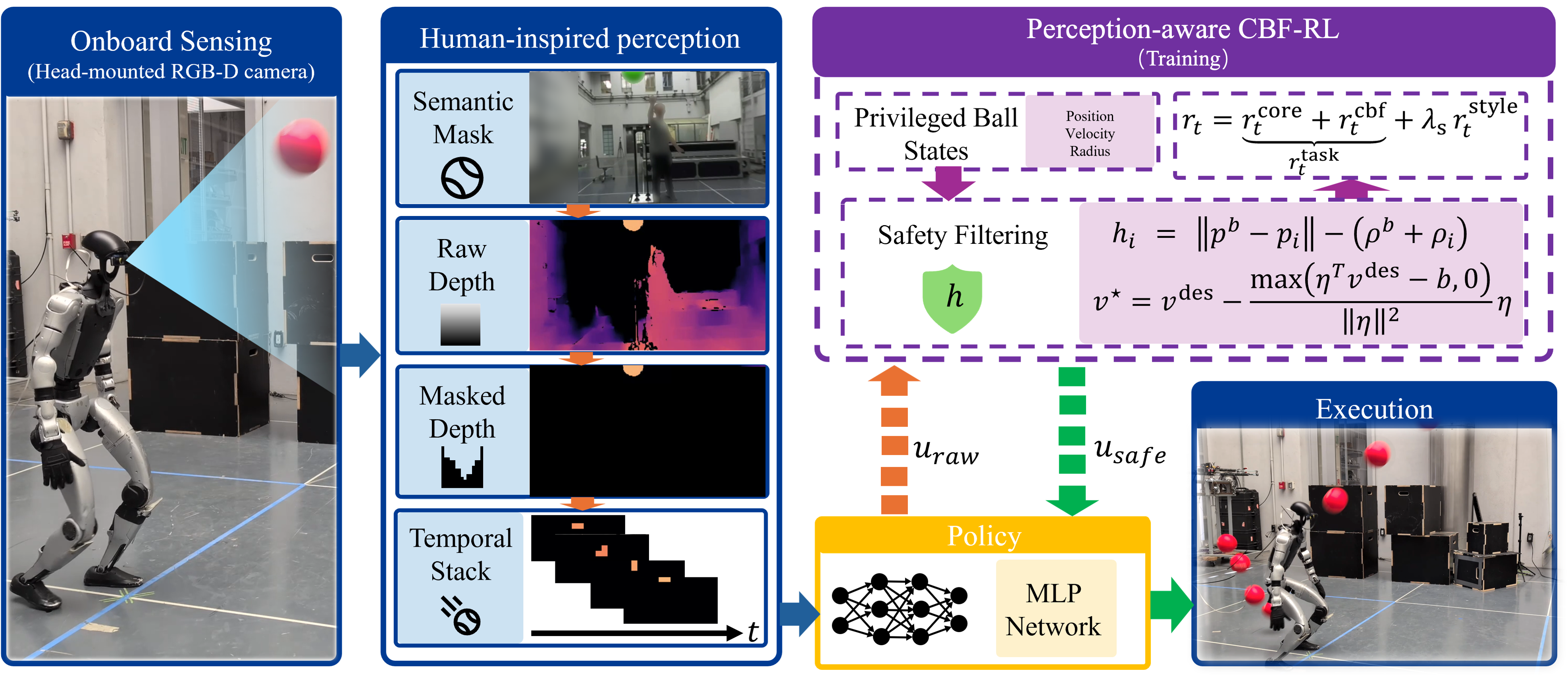}
  \caption{Pipeline overview.
  The policy maps temporally stacked ball-only depth and proprioception to $29$-D joint targets; during training, ball-to-link geometry supplies whole-body CBF guidance that the policy internalizes for deployment.}
  \label{fig:system-overview}
\end{figure*}
\subsection{Related Work}
\label{sec:relatedwork}

\newsec{Motion imitation and priors for humanoid control.}
Motion imitation has long provided a practical route to naturalistic, high-dimensional control.
\cite{peng2018deepmimic} showed that RL can track reference motions reliably, \cite{peng2021amp} replaced explicit tracking with an adversarial discriminator over motion features, which yields a style reward compatible with a task objective, and \cite{peng2022ase,mu2026smp} learned reusable skill and style embeddings over large motion datasets.
These build on a broader body of legged learning \cite{li2021reinforcement,boxlocomanip2023,dai2026walk} trained from task rewards, command tracking, and domain randomization.
Recent whole-body tracking systems move closer to our humanoid setting:
\cite{amo2025} investigates dexterous whole-body control, \cite{opt2skill2024} uses optimized trajectories to train policies, and \cite{beyondmimic2025} constructs a standardized whole-body imitation training pipeline.
Retargeting quality also matters here, both for general humanoid tracking \cite{araujo2025retargeting} and for interaction-preserving loco-manipulation \cite{omniretarget2025}.
For dodgeball, a motion prior is attractive because the reward for ``not being hit'' is too semantically sparse and geometric to specify how a humanoid should move.
We select human dodging motions as priors and also condition the policy on a perception stream, thus combining the motion prior with exteroceptive information to produce whole-body evasive reflexes.

\newsec{Perception-aware safety, parkour, and fast-object tasks.}
The closest conceptual neighbor to perception-aware dodgeball is safe agility under partial exteroception.
Quadruped parkour \cite{cheng2023extreme, hoeller2024anymal} and agile-but-safe locomotion \cite{he2024agilebut} show how perception, skill selection, recovery, and safety layers can support fast motion around obstacles, while humanoid parkour and perception-aware locomotion systems \cite{long2024pim,zhuang2025humanoidparkour,php2026,zhuang2026deepwholebody} extend this to higher-dimensional bodies with depth- or elevation-conditioned policies for jumping, leaping, and terrain traversal.
Related work studies the same perception–speed–stability tradeoff through depth conditioning \cite{hao2026cref}, online policy switching \cite{baek2026bat}, and active gaze \cite{li2026taga}.
Dodgeball differs from these terrain-centered settings because the hazard is a fast-moving object rather than traversable terrain.
The ball may be visible only briefly, so the policy must forecast imminent body contact and produce an evasive whole-body response instead of merely choosing footholds.
Task-centered fast-object work makes this timing pressure explicit: \cite{hitter2025} uses humanoid table tennis to study sub-second ball prediction, task planning, and whole-body RL control.
Dodgeball shares the timing but changes the objective from striking to avoiding contact.
This shift stresses egocentric observability: an onboard depth camera must provide enough information for the body to leave the ball's path before impact.

\newsec{Whole-body avoidance and safety.}
Reactive safety spans both recovering balance after a disturbance via capture-point and push-recovery control \cite{pratt2006capture,stephens2010dynamic}, and avoiding contact in the first place via control barrier functions \cite{ames2016control}, collision-free MPC \cite{chiu2022collision}, and fast collision-free motion generation \cite{sundaralingam2023curobo}.
Most related to our safety treatment, \cite{yang2025cbf} injects safety structure during humanoid training and removes the runtime filter at deployment, and \cite{kohlbrenner2026egocentric} concurrently explores egocentric tactile and proximity sensing as a modality complementary to the vision used here.
We study depth-based dodgeball on the Unitree G1, using perception-aware CBF-RL to couple onboard visual observations with per-link safety guidance and motion-regularized whole-body evasion.

\section{Safety-Critical Perception-aware Dodgeball}
\label{sec:problem}

We formulate perception-aware dodgeball as a safety-critical whole-body reaction problem under partial observation.
The robot must avoid contact with an incoming object while preserving balance, yet its deployed policy has only limited onboard sensing.
This information constraint affects both which safety mechanisms we can use and how we measure success.

\subsection{Task and observations}
\label{sec:m-formulation}

A ball is launched toward a standing humanoid, which must avoid being struck while preserving balance.
We model the task as a partially observed Markov decision process.
At control step $t$, the policy $\pi_\theta(a_t \mid o_t)$ emits joint-position targets $a_t \in \mathbb{R}^{29}$ (offsets from a nominal pose, tracked by PD actuators at 50\,Hz).

The defining constraint is deployability: the policy observation $o_t$ contains only signals computable on hardware from onboard sensing,
\begin{equation}
  o_t = \big(\, \mathcal{D}_t,\; q_t,\dot q_t,\; \omega_t,\; g_t,\; a_{t-1} \,\big),
  \label{eq:policy-obs}
\end{equation}
where $\mathcal{D}_t$ is the depth observation, $q_t,\dot q_t$ are joint positions/velocities, $\omega_t$ is the base angular velocity, $g_t$ is the projected gravity, and $a_{t-1}$ is the previous action.
For the camera-based policies, the ball's state is never part of $o_t$; whatever the policy knows about the ball, it must infer from depth.

\subsection{Projectile threat model}
\label{sec:m-threat}

Balls are thrown on an intermittent timer that leaves a recovery window between throws.
The default distribution launches from a frontal cone (within $\pm 25^\circ$ of heading, 2-3\,m ahead) with a flight time of roughly $0.6$\,s, which sets the reaction window; the aim point leads the robot's velocity.
We mix two threat types equally so the policy learns both evasions: a descending ball that falls across the legs (sidestep/step-over) and a low-arc ball that rises to torso/head height (duck/lean).
Throw heights are calibrated against a frozen-statue control (\S\ref{sec:Results}) so that essentially every throw is on-target.
We partition environments at reset into $20\%$ ball-free standing anchors and $80\%$ ball-thrown evade episodes.

\subsection{Whole-body safety metric}
\label{sec:m-success}

The robot is hit if the ball contacts any body link and falls if the torso orientation, base height, or sustained crouch crosses a threshold.
A trial succeeds if neither event occurs within the reaction window.
The benchmark reports reset and deployment regimes in \S\ref{sec:Results}; reset success is the primary single-throw metric.
This contact-based score is intentionally stricter than measuring the distance from the ball to the pelvis alone: many failures are limb grazes, so a method that only protects the base can appear safe while still being struck.

\section{Methods}
\label{sec:methods}

\subsection{Safety-guided learning}
\label{sec:m-rl}

PAC-MAN trains a perception-aware policy to map partial onboard observations to whole-body actions shaped by barrier information available only during training.
Privileged ball and link states instantiate the barrier conditions, while the policy receives only the onboard observation $o_t$ of Eq.~\eqref{eq:policy-obs}.
We use CBF-RL to inject violations of these conditions into policy optimization~\cite{yang2025cbf}, so the policy internalizes the safety structure during training and needs no privileged geometry at deployment.
Link-CBF is a lightweight per-link reward that extends the pelvis-centered task objective to every body link, while Joint-CBF is a stronger joint-space projection that can shape training or remain active as a privileged runtime filter.
Segmentation-masked depth is the threat representation we deploy, and an adversarial motion prior regularizes the resulting reactions without defining the safe set.

Figure~\ref{fig:system-overview} shows the full pipeline.
The full reward is
\begin{equation}
  r_t \;=\; \underbrace{r^{\mathrm{core}}_t \;+\; r^{\mathrm{cbf}}_t}_{r^{\mathrm{task}}_t} \;+\; \lambda_{\mathrm{s}}\, r^{\mathrm{style}}_t,
  \label{eq:reward}
\end{equation}
where the task reward pairs the distance-to-core evasion term $r^{\mathrm{core}}$ of prior humanoid dodgeball work~\cite{mu2026smp} with the control-barrier term $r^{\mathrm{cbf}}$.
The reward also contains station-hold, posture, and smoothness terms that we omit from the notation, and $r^{\mathrm{style}}$ regularizes the form of the motion (\S\ref{sec:m-motion}).
Table~\ref{tab:reward} gives the full reward stack with formulations and weights.
The distance-to-core term rewards keeping the ball far from the pelvis while otherwise standing still,
\begin{equation}
  r^{\mathrm{core}}_t \;=\; w_{\mathrm{pos}} \Big(1 - e^{-k_{\mathrm{pos}} \lVert p^{b}_t - p^{\mathrm{pel}}_t \rVert}\Big) \;+\; w_{\mathrm{vel}}\, e^{-k_{\mathrm{vel}} \lVert v^{\mathrm{pel}}_{xy,t} \rVert^{2}},
  \label{eq:core}
\end{equation}
where $p^{b}, p^{\mathrm{pel}}$ are the ball and pelvis positions and $v^{\mathrm{pel}}_{xy}$ is the horizontal base velocity.
The distance term saturates once the ball is far, so the robot settles rather than flees, and the stillness term makes it move only when it must evade.
The CBF term is the safety axis we vary in \S\ref{sec:Results}: the no-barrier baseline sets $r^{\mathrm{cbf}}=0$, and the Link-CBF and Joint-CBF forms of \S\ref{sec:m-cbf} add increasing barrier structure.
The hit termination is active from the first training iteration and carries no explicit penalty: the only cost of being hit is the future reward foregone by an early episode end.
Training follows the asymmetric actor-critic paradigm: the value function $V_\phi(s_t)$ observes a privileged state $s_t$ that augments the policy observation $o_t$ of Eq.~\eqref{eq:policy-obs} with the ground-truth ball relative position/velocity, radius, a visibility gate, and full-body link kinematics.
These privileged signals support training and barrier computation but are discarded at deployment, so the policy must recover the relevant threat information from $o_t$ alone.

\subsection{Whole-body barriers}
\label{sec:m-cbf}

\begin{figure}[t!]
\vspace{0.2cm}
  \centering
  \includegraphics[width=\linewidth]{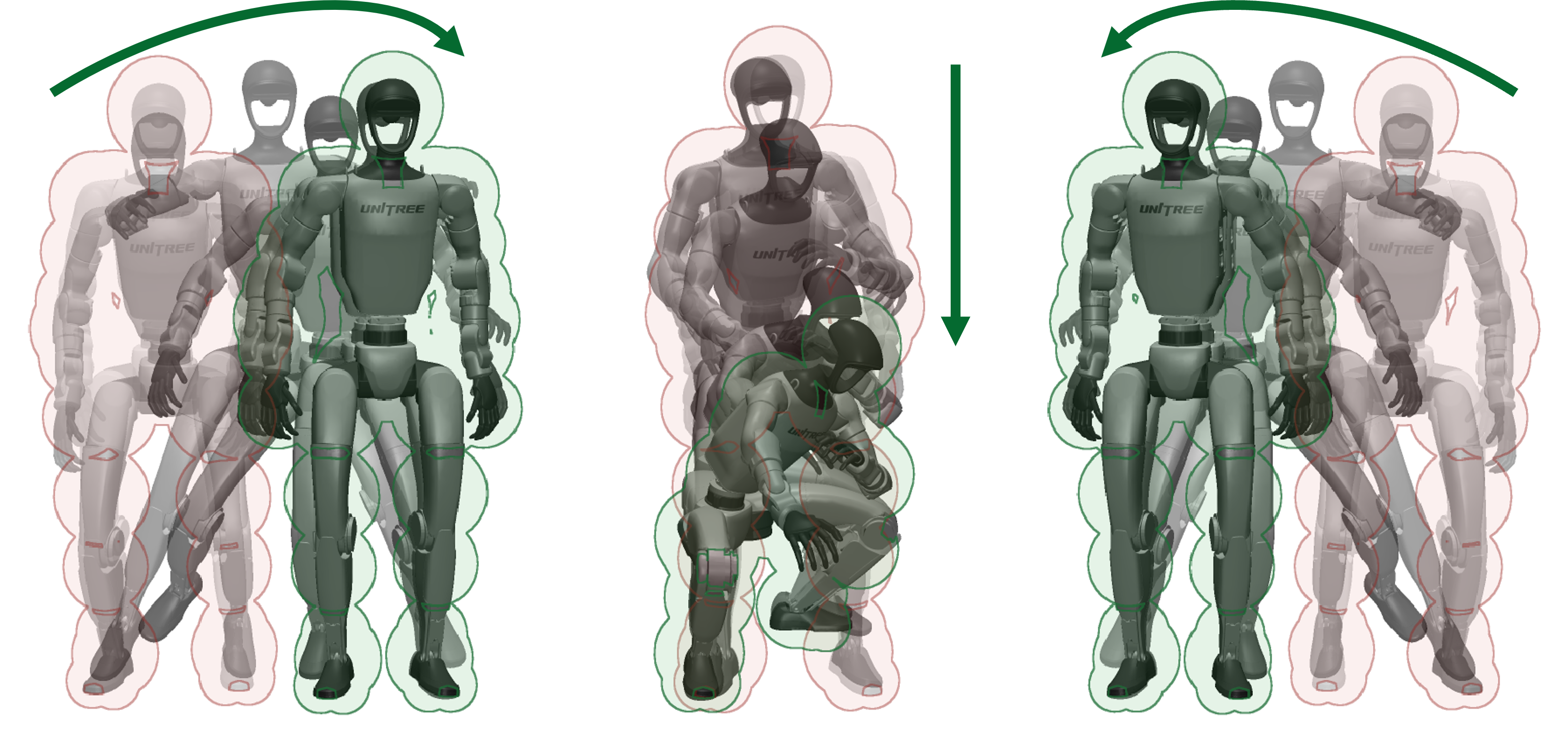}
  \caption{Whole-body barrier geometry across representative motion-prior clips; the outlines show the union of per-link keep-out shells as the humanoid crouches, leaps, and sidesteps.
  The reference set contains 1 minute and 43 seconds of human motion retargeted to the Unitree G1 using \cite{bones_seed_2026}.}
  \label{fig:barrier_motion}
\end{figure}

Control barrier functions encode a safe set through a function whose nonnegative superlevel set should remain invariant under an admissible controller~\cite{ames2016control}.
PAC-MAN uses barrier information at two levels to study what a policy can internalize from partial perception and what still requires direct online enforcement.
The lightest form sets the safety term of Eq.~\eqref{eq:reward} to a per-link clearance reward, $r^{\mathrm{cbf}} = \lambda_{\mathrm{cbf}}\, r^{\mathrm{clear}}$.
We form the clearance for every robot link $i$,
\begin{equation}
  h_i \;=\; \lVert p^{b} - p_i \rVert - \big(\rho^{b} + \rho_i\big),
  \label{eq:linkcbf-h}
\end{equation}
where $p^{b}, p_i$ are the ball and link positions and $\rho^{b}, \rho_i$ their effective radii (collision cross-section plus a small margin), and its closing rate $\dot h_i$ follows from the relative ball-link velocity.
The corresponding whole-body collision-free set is $\mathcal{C}=\bigcap_i\{\,h_i \ge 0\,\}$, visualized in Fig. \ref{fig:barrier_motion}, so protecting only the pelvis is insufficient whenever the ball threatens an arm, leg, or torso link.
The reward penalizes only violations of the barrier condition $\dot h_i + \alpha h_i \ge 0$ at the most-binding link:
\begin{equation}
  r^{\mathrm{clear}} \;=\; \min_i\, \mathrm{clip}\big(\dot h_i + \alpha h_i,\; -c,\; 0\big),
  \label{eq:linkcbf}
\end{equation}
gated to zero unless a thrown ball is airborne and moving, so the nominal stand is untouched.
The class-$\mathcal{K}$ term $\alpha h_i$ ramps the penalty up over the final approach rather than only at contact, and the clip $c$ keeps the term from acting like a hard terminal penalty.
This lightweight per-link barrier, which we call Link-CBF, represents threats to limbs and other links that the pelvis-centered $r^{\mathrm{core}}$ does not see.
Because Link-CBF enters only through the training reward, it guides the policy toward the safe set but does not enforce the barrier condition online during deployment.

The Joint-CBF condition adds a stronger whole-body joint-space CBF module during training.
The policy's joint-position target $q^{\mathrm{des}}$ (the nominal pose plus $a_t$) defines an implicit desired joint velocity $v^{\mathrm{des}} = (q^{\mathrm{des}} - q_t)/\Delta t$.
The module defends every body point with the squared-distance barrier $h_i = \lVert p^{b} - p_i \rVert^{2} - D_i^{2}$, where $D_i = \rho^{b} + \rho_i + d_{\mathrm{buf}}$ is a keep-out radius, and at each step selects the most-threatened point $i^{\star} = \arg\min_i h_i$.
Enforcing $\dot h_{i^{\star}} \ge -\alpha h_{i^{\star}} + \delta$ then yields a single affine constraint on the joint velocity,
\begin{equation}
  \eta^{\top} v \;\le\; b,
  \qquad
  \eta = 2\, J_{i^{\star}}^{\top} r,
  \quad
  b = \alpha h_{i^{\star}} + 2\, r^{\top} \dot p^{\,b} - \delta,
  \label{eq:jointcbf-con}
\end{equation}
where $r = p^{b} - p_{i^{\star}}$, $J_{i^{\star}}$ is the point's positional Jacobian, and $\delta = \delta_{\max}\,[\,1 - T/T_{\mathrm{alert}}\,]_{+}$ is an urgency margin that tightens the constraint as the time to contact $T$ drops below $T_{\mathrm{alert}}$.
The barrier-projected velocity is the closed-form projection of $v^{\mathrm{des}}$ onto this half-space,
\begin{equation}
  v^{\star} \;=\; v^{\mathrm{des}} \;-\; \frac{\big[\,\eta^{\top} v^{\mathrm{des}} - b\,\big]_{+}}{\lVert \eta \rVert^{2}}\; \eta,
  \label{eq:jointcbf-proj}
\end{equation}
which equals $v^{\mathrm{des}}$ whenever the constraint already holds or no threat is sensed.
Joint-CBF adds two shaping terms to $r^{\mathrm{cbf}}$: a correction cost $-\lambda_{\mathrm{corr}}\, \lVert v^{\star} - v^{\mathrm{des}} \rVert^{2}$ that penalizes how hard the filter has to intervene and a buffer cost $-\lambda_{\mathrm{buf}}\, \big[\, d_{\mathrm{safe}} - h_{\min} \,\big]_{+}^{2}$ on the linear clearance $h_{\min}$ of the most-threatened point, which discourages skimming the keep-out boundary.
At evaluation, we remove this training-time module, so the policy must execute its own learned avoidance from depth and proprioception alone.
This reward-only Joint-CBF condition measures how much of the stronger barrier structure the perception-aware policy can internalize.
We also report a privileged ceiling, denoted +filter, that keeps the projection of Eq.~\eqref{eq:jointcbf-proj} active at test time and executes $q_t + v^{\star} \Delta t$ in place of the policy target.
The filter reads ground-truth ball states online, so it is not a deployable vision-only configuration; it measures what direct barrier enforcement achieves when accurate threat state is available.

\subsection{Perception}
\label{sec:m-perception}

The barriers above are computed from ball and link geometry during training, but the deployed policy must infer the relevant threat geometry from its observations.
Perception, therefore, controls how faithfully the policy can associate a visual threat with the barrier signal used to train it.
We isolate this informational effect with three regimes that change the ball representation but leave the underlying control task unchanged:

\newsec{State oracle.} A symmetric actor-critic that observes the ground-truth ball states directly (no camera).
This is the safety-information upper bound.

\newsec{Fixed depth camera.} The focusing stage alone: the ball is segmented from the RGB-D stream of a single rigidly mounted head camera ($102^\circ$ horizontal FoV, $57^\circ$ vertical FoV, tilted $20^\circ$ up).
In simulation, we build this observation from the simulator's segmentation: pixels on the ball retain their measured depth, every other pixel is set to the far plane, and the masked image is pooled to $16\times9$.
We stack the masked frame with older snapshots at sparse temporal offsets rather than consecutive frames (specifically the last $[0,3,8,18]$ frames with the camera stream at 50 Hz): they are far enough apart to encode looming and lateral motion while keeping the observation small enough for a feed-forward policy.
The camera is rigidly mounted, so a fast ball that leaves its field of view or shrinks below the angular resolution at range is unobservable.
During training, we randomize the masked observation with whole-ball dropout, ragged pixel dropout, depth jitter, and edge flicker to model failures of the real perception stack.

\newsec{Active gimbal (gaze).} The second part of human-like perception added on top of the focusing stage: we mount the same camera on a $1$-DoF pitch joint ($\pm 30^\circ$, nominal position at $0^\circ$) so that it can track the ball through its flight, in analogy to human gaze, which orients toward a threat to keep it foveated.
The robot also receives the aim angle as proprioception.
We establish the ceiling of this design by driving the joint with an oracle that points the optical axis at the ball; the policy still dodges from depth alone, but the camera is perfectly aimed.
This isolates the observability effect from the control problem: the policy receives the same low-resolution depth representation either way, but the ball remains in view for more of its flight.

\newsec{Toward deployable active perception.}
The oracle-aimed gimbal requires the ground-truth ball states to aim, so it is a simulation ceiling.
Closing the loop does not require the policy to learn the aim: a deployable version would drive the same pitch joint from the segmentation track already computed by the focusing stage, pointing the camera at the tracked ball rather than exposing the aim as a policy action.
Realizing this tracker-aimed gimbal on the physical robot is future work, since it requires additional hardware; the system we deploy uses the fixed camera.

\subsection{Motion prior and training}
\label{sec:m-motion}

\begin{table}[t]
\vspace{0.2cm}
\centering
\caption{Reward terms of the dodge task with formulations and training weights. $\mathbb{I}[\mathrm{thr}]$ is the throw indicator function.}
\label{tab:reward}
\setlength{\tabcolsep}{3pt}
\small
\resizebox{\columnwidth}{!}{%
\begin{tabular}{l l r}
\toprule
Term & Expression & Weight \\
\midrule
\multicolumn{3}{l}{\textit{Task, safety, and style}} \\
Distance-to-core $r^{\mathrm{core}}$ & \makecell[l]{$0.9\big(1-e^{-0.3 \lVert p^{b}-p^{\mathrm{pel}}\rVert}\big)$\\$\quad+\, 0.1\, e^{-\lVert v^{\mathrm{pel}}_{xy}\rVert^{2}}$} & $1.0$ \\
Link-CBF $r^{\mathrm{clear}}$ & \makecell[l]{$\min_i \mathrm{clip}(\dot h_i {+} \alpha h_i, -c, 0)$\\$\quad\alpha{=}1$, $c{=}2$} & $0.27$ \\
Joint-CBF correction & $-\lVert v^{\star}-v^{\mathrm{des}}\rVert^{2}/n_{j}$ & $0.1$ \\
Joint-CBF buffer & \makecell[l]{$-\mathbb{I}[\mathrm{thr}]\, \big[\, d_{\mathrm{safe}} - h_{\min} \,\big]_{+}^{2}$\\$\quad d_{\mathrm{safe}}{=}0.1$} & $1.0$ \\
AMP style $r^{\mathrm{style}}$ & $\max\big[0,\, 1-\tfrac{1}{4}\big(D_\psi-1\big)^{2}\big]$ & $0.5$ \\
\midrule
\multicolumn{3}{l}{\textit{Station hold and posture}} \\
Station hold & $e^{-d_{\mathrm{home}}^{2}/1.5^{2}}$ & $0.3$ \\
Stillness when safe & $(1{-}\mathbb{I}[\mathrm{thr}])\, e^{-2\lVert v^{\mathrm{pel}}_{xy}\rVert^{2}}$ & $0.5$ \\
Action rate when safe & $(1{-}\mathbb{I}[\mathrm{thr}])\, \lVert a_t {-} a_{t-1}\rVert^{2}$ & $-0.05$ \\
Base height & $e^{-(z^{\mathrm{des}} - z^{\mathrm{pel}})^{2}/0.3^{2}}$ & $1.0$ \\
Angular stability & $e^{-\lVert \omega_{xy}\rVert^{2}/\pi^{2}}$ & $0.5$ \\
\midrule
\multicolumn{3}{l}{\textit{Regularizers and terminations}} \\
Fall termination & $\mathbb{I}[\mathrm{fall}]$ & $-200$ \\
Joint acceleration & $\lVert \ddot q \rVert^{2}$ & $-2.5\times10^{-7}$ \\
Joint limit violation & $\textstyle\sum_j \big[\, |q_j| - q_j^{\mathrm{lim}} \,\big]_{+}$ & $-10$ \\
Action rate & $\lVert a_t - a_{t-1}\rVert^{2}$ & $-0.01$ \\
Foot slip & $\textstyle\sum_{f \in \mathrm{contact}} \lVert v^{f}_{xy}\rVert^{2}$ & $-0.25$ \\
Self-collision & contacts $>10$\,N & $-0.1$ \\
\bottomrule
\end{tabular}}
\end{table}

The CBF terms specify the safety structure of the task, but they do not prescribe which dynamically feasible whole-body motion should realize an evasion.
We therefore add an adversarial motion prior (AMP)~\cite{peng2021amp} as a motion regularizer.
A discriminator $D_\psi$ is trained to distinguish policy state transitions from transitions drawn from a reference set $\mathcal{M}$ of human-derived whole-body clips\footnote{Motion Data by Bones Studio - https://bones.studio/} (crouching, leaning, sidestepping, leaping, and standing), and the policy receives the style reward
\begin{equation}
  r^{\mathrm{style}}_t \;=\; \max\!\Big[\,0,\; 1 - \tfrac{1}{4}\big(D_\psi(s_t,s_{t+1}) - 1\big)^2\Big],
  \label{eq:amp}
\end{equation}
which favors reflex-like crouches, leans, and sidesteps over arbitrary joint excursions.
Training uses PPO and the AMP discriminator from \texttt{rsl\_rl}~\cite{schwarke2025rsl}, with $8192$ parallel environments for $20$k iterations in mjlab~\cite{zakka2026mjlab,amp_mjlab}.
The depth feed is capped at $16\times9$ resolution.
The full reward construction is shown in Table~\ref{tab:reward}.

\section{Results}
\label{sec:Results}

\begin{figure*}[t]
  \centering
  \includegraphics[width=0.9\textwidth]{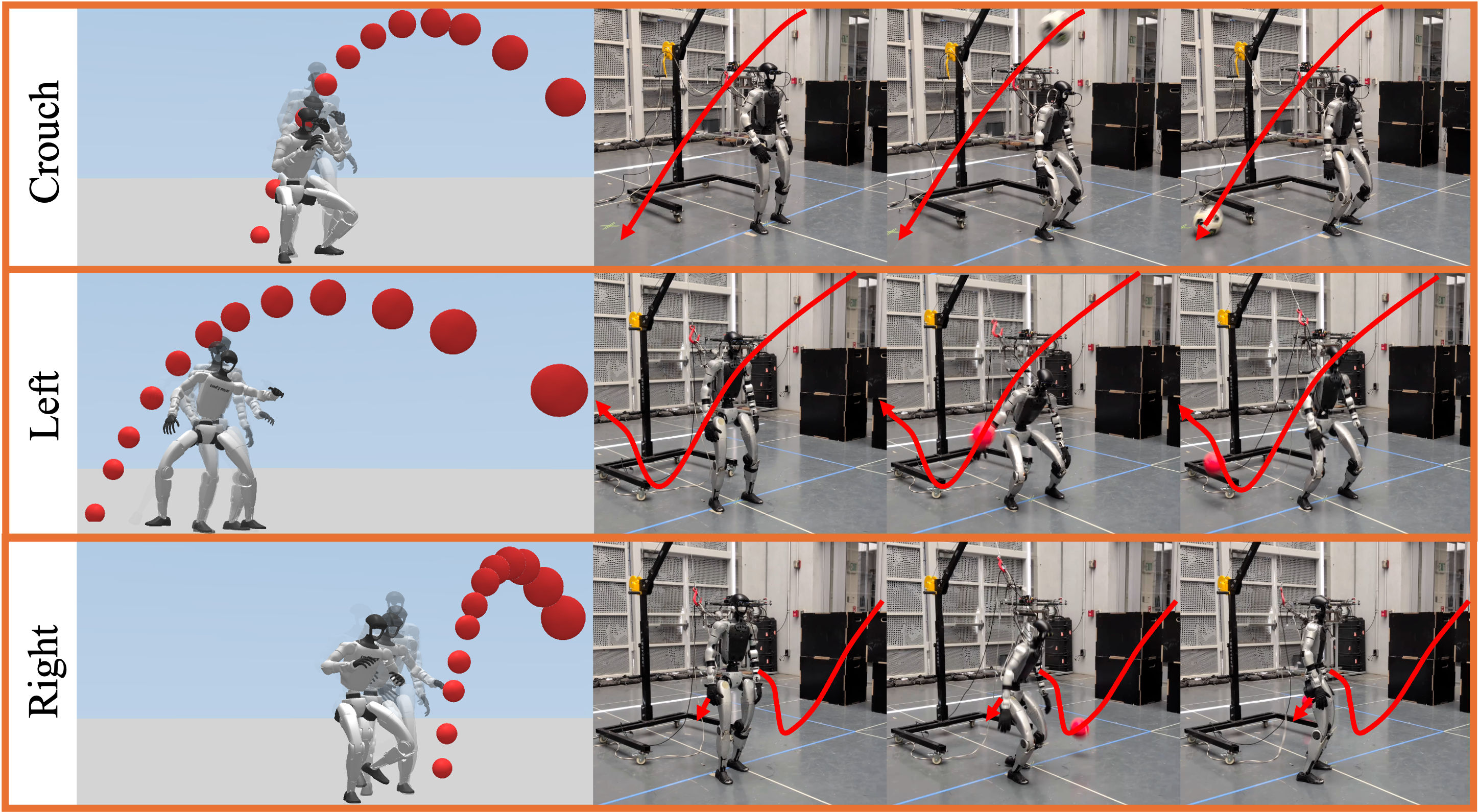}
  \caption{Emergent whole-body evasion modes in simulation (left) and hardware (right).
  Crouching, leaning, and sidestepping coordinate the legs, torso, and arms to move threatened links rather than only translating the base.
  Semantic segmentation allows the same policy to dodge different balls, shown with a soccer ball (top row) and a foam ball (lower two rows).}
  \label{fig:action-montage}
  \vspace{-5pt}
\end{figure*}

We first evaluate the PAC-MAN framework in simulation to quantify the impact of perception and safety structure; then we verify the full stack on the physical robot.
Figure~\ref{fig:action-montage} shows emergent evasion modes from both.

\subsection{Quantitative evaluation in simulation}
\label{sec:sim}
In simulation, we compare the onboard fixed-camera policy against a gimbal-mounted camera and a privileged state oracle while varying the safety structure at each perception level, from no barrier to the full joint-space CBF with its runtime filter.
All policies are evaluated on the same set of seeded throws and scored by the environment's true terminations: success requires avoiding contact with every body link and remaining upright, rather than relying on pelvis clearance alone.
A static robot survives only $4\%$ of throws, which sets the on-target floor.
We evaluate every policy at a matched $20$k-iteration checkpoint.
We report two throw regimes.
\textbf{Reset}: We reset the robot between throws, and the robot dodges immediately.
\textbf{Deployment}: A locomotion controller returns the robot to the origin between throws before the next dodge.
We vary the safety reward along four modes:
\begin{itemize}
  \item \textbf{No barrier} uses the distance-to-core reward $r^{\mathrm{core}}$ only ($r^{\mathrm{cbf}} = 0$).
  \item \textbf{Link-CBF} adds the lightweight per-link clearance barrier $r^{\mathrm{clear}}$ as reward guidance (\S\ref{sec:m-cbf}); this is the configuration we deploy on hardware.
  \item \textbf{Joint-CBF} additionally trains with the whole-body joint-space CBF module as reward guidance, evaluated without a runtime filter.
  \item \textbf{+filter} keeps the joint-space CBF as an online filter.\footnote{The filter reads ground-truth ball states, so running it on hardware would require a highly accurate online ball state estimator that our vision stack does not provide.}
\end{itemize}

\begin{table}[t]
\centering
\includegraphics[width=\columnwidth]{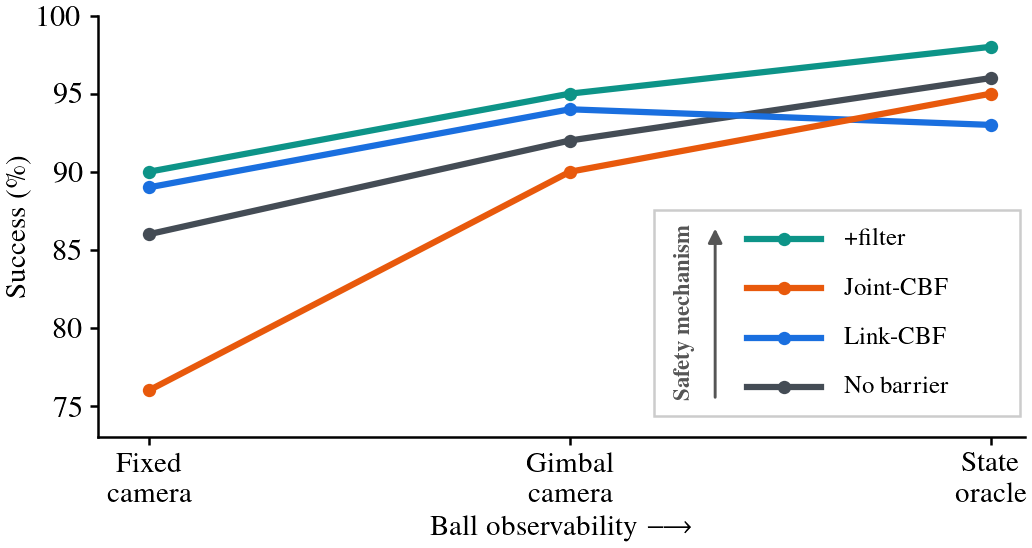}\\[0.5em]
\setlength{\tabcolsep}{3.5pt}
\resizebox{\columnwidth}{!}{%
\begin{tabular}{ll cccc c}
\toprule
Regime & Perception & No barrier & Link-CBF & Joint-CBF & +filter & Fall \\
\midrule
\multirow{3}{*}{Reset}
 & State oracle   & 98 & 88 & 96 & 99 & 0 \\
 & Fixed camera   & 86 & \textbf{90} & 83 & 97 & 0 \\
 & Gimbal camera  & 96 & 95 & 91 & 97 & 0 \\
\midrule
\multirow{3}{*}{Deployment}
 & State oracle   & 96 & 93 & 95 & 98 & 0 \\
 & Fixed camera   & 86 & \textbf{89} & 76 & 90 & 0 \\
 & Gimbal camera  & 92 & 94 & 90 & 95 & 0 \\
\bottomrule
\end{tabular}}
\caption{Whole-body success (no link contact or fall) and fall rates (\%) by throw regime, perception, and safety-reward mode (defined in \S\ref{sec:sim}) at a matched $20$k-iteration checkpoint.
Bold marks the deployed fixed-camera Link-CBF cells.
Top: deployment success versus ball observability, one line per safety mode; Joint-CBF improves sharply with observability, Link-CBF remains consistent, and +filter requires runtime ball states.}
\label{tab:main}
\end{table}

Onboard sensing costs only a few points of evasion.
Trained with the same Link-CBF reward, the fixed-camera policy ($90\%$ reset, $89\%$ deployment) is on par with the state oracle ($88\%$ and $93\%$) despite seeing the ball only as a conservative masked depth (Table~\ref{tab:main}).
The oracle's best configurations ($96$-$99\%$) sit fewer than ten points above the policy we deploy.
The fixed onboard camera alone is therefore adequate for evasion; what perception changes is which safety structure works.

With privileged ball states, the strongest safety configuration achieves the highest success rates.
The plain distance-to-core baseline is already strong ($96\%$-$98\%$ across regimes), and the full CBF-RL recipe, with the joint-space barrier kept as its runtime filter, outperforms it ($98\%$-$99\%$).
The +filter column is the best mode in every row of Table~\ref{tab:main}: given accurate online ball states, runtime whole-body barrier enforcement achieves the highest success rate at each perception level.
This improvement requires ball states that our onboard stack currently cannot supply.

Among the modes that run without runtime ball states, the ranking flips.
Link-CBF is best on the fixed camera in both regimes (reset $90\%$, deployment $89\%$), above the no-barrier baseline ($86\%$ in both) and well above the bare Joint-CBF, which drops to $76\%$ in the deployment regime.
Link-CBF extends safety guidance from pelvis clearance to every body link; under the any-link contact criterion, it improves fixed-camera success from $86\%$ to $90\%$ in reset and $89\%$ in deployment.
It is therefore the reward we deploy on hardware (\S\ref{sec:hw}).

We theorize that the Joint-CBF gap comes from perception: the policy is asked to internalize a whole-body barrier that it cannot perceive accurately enough, and it learns worse evasions instead.
Two changes repair this, and both supply the ball information that the fixed camera lacks.
Keeping the joint-space CBF as an online filter hands exact ball states to the barrier itself and lifts the fixed camera to $90\%$-$97\%$.
Giving the policy a better view works as well: with an oracle-aimed, ball-tracking gimbal, the same Joint-CBF training no longer hurts (deployment $90\%$ versus $76\%$ on the fixed camera) and beats its no-barrier counterpart with the full safety structure (i.e. +filter).
This shows that the whole-body barrier succeeds only when the ball is sufficiently observable.
Fig.~\ref{tab:main} (top) illustrates this: deployment success of Joint-CBF climbs steeply with ball observability, while Link-CBF stays nearly flat.
Therefore, the design implication is that the strength of the safety structure must be matched to the information available.
A hardware gimbal aimed by the same segmentation-based focusing stage that already feeds the policy is left as future work.
We also compare against prior results reported in SMP~\cite{mu2026smp} with an omnidirectional throw distribution ($360^\circ$, $8$-$10$\,m, $12$-$15$\,m/s in codebase) with oracle observations.
Under this matched setting, PAC-MAN scores $98$\% reset, in line with their reported $\sim$99\%.

\subsection{Hardware experiments}
\label{sec:hw}

We deploy the Link-CBF fixed-camera policy on the Unitree G1 with a head-mounted ZED Mini.
On hardware, the perception node replaces the simulator's segmentation with a learned tracker: an EfficientTAM track, initialized from a clicked foreground point, segments the ball in RGB at ${\sim}60$\,Hz, above the $50$\,Hz control rate, so the masked-depth observation is effectively fresh with each control step.
This also gives us the ability to dodge different balls since the tracker can be initialized on any object of interest.
An example of dodging a soccer ball instead of a foam ball is shown in the top row of Fig.~\ref{fig:action-montage}.
The tracked object's ZED depth pixels are retained, and all others are set to far range before pooling to the same $16\times9$ stacked-depth observation used in training.
Several filters keep this feed conservative: low-percentile pooling preserves the ball while rejecting isolated ZED flying pixels; distance-aware mask-size checks reject masks that are too tall to be a dodgeball at their measured range; and an optional looming gate suppresses tracked clusters whose average depth is not closing quickly enough, preventing a static object or over-sensitive segmentation from triggering dodges.
If the track is absent, rejected by these checks, or lost, the policy receives an all-far frame rather than a false incoming ball.
Proprioception comes from the robot's IMU and encoders and uses a walk$\leftrightarrow$dodge mode switch to return the robot to a fixed location between throws.
No ground-truth ball states, velocity estimates, or runtime CBF inputs are available at any point: the policy reacts purely to the conservative masked-depth feed, exactly as in the fixed-camera Link-CBF cells of Table~\ref{tab:main}.
We stress-test this stack by throwing balls at the robot from the front. 
A throw counts as dodged only if the ball contacts no body link.
The policy dodges $19$ of $20$ throws ($95\%$, Table~\ref{tab:hw}).
Fig.~\ref{tab:hw} (top) shows the estimated barrier value $h$ for each throw.
The hit is shown at contact, and the remaining throws clear the per-link keep-out shell by $0.05$ to $1.7$\,m.
The sim-trained policy therefore transfers to the physical robot from onboard depth alone.

\begin{table}[h]
\centering
\includegraphics[width=\columnwidth]{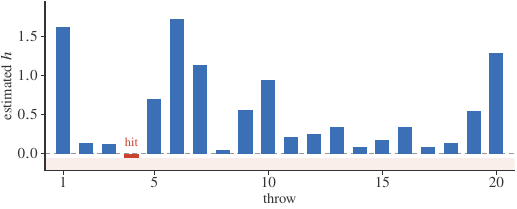}\\[0.5em]
\begin{tabular}{l c c c c}
\toprule
Throws & Dodged & Hit & Falls & Success (\%) \\
\midrule
20 & 19 & 1 & 0 & 95 \\
\bottomrule
\end{tabular}
\caption{Hardware dodge outcome: $20$ balls thrown by hand at the deployed Link-CBF fixed-camera policy on the Unitree~G1 (policy input: onboard depth; RGB used only for segmentation), with contact on any body link counted as a hit.
Top: estimated barrier value $h$ for each throw; the dashed line is the per-link keep-out shell ($h=0$), and the hit (throw 4, red) is shown at contact.}
\label{tab:hw}
\end{table}

\section{Conclusion and Future Work}
\label{sec:Conclusion}

We presented PAC-MAN, a perception-aware CBF-RL framework for whole-body humanoid dodgeball, and deployed its fixed-camera Link-CBF policy on the Unitree G1 using only onboard depth and proprioception.
PAC-MAN couples segmentation-masked depth with control-barrier guidance computed during training, and a human motion prior regularizes the form of the evasive reflexes.
Our results show that perception-aware CBF-RL is a safety-information co-design problem: the appropriate barrier structure depends on the threat representation available to the policy or filter.
With accurate ball states, the Joint-CBF runtime filter is the strongest configuration, whereas the same joint-space barrier used only as reward guidance degrades the fixed-camera policy.
Link-CBF extends the pelvis-centered objective to every body link while tolerating limited onboard perception, and the deployed policy succeeds on $95\%$ of physical throws without privileged ball states or a runtime filter.
The segmentation front end also lets one policy evade different balls.
These results distinguish internalizing barrier structure in a learned policy from enforcing a barrier online: internalization is limited by what the policy can infer, while enforcement requires accurate threat state at runtime.
Future work will close this perception-to-CBF loop with a physical tracker-aimed gimbal or an accurate online ball-state estimator and extend the task from dodging at a recovered stance to dodging while walking.

\balance
\begingroup
\scriptsize
\bibliographystyle{ieeetr}
\bibliography{reference.bib}
\endgroup

\end{document}